# Exploration Of The Dendritic Cell Algorithm Using The Duration Calculus

Feng Gu, Julie Greensmith and Uwe Aickelin

School of Computer Science, University of Nottingham
Nottingham, NG8 1BB, UK
{fxg,jqg,uxa}@cs.nott.ac.uk

**Abstract.** As one of the newest members in Artificial Immune Systems (AIS), the Dendritic Cell Algorithm (DCA) has been applied to a range of problems. These applications mainly belong to the field of anomaly detection. However, real-time detection, a new challenge to anomaly detection, requires improvement on the real-time capability of the DCA. To assess such capability, formal methods in the research of real-time systems can be employed. The findings of the assessment can provide guideline for the future development of the algorithm. Therefore, in this paper we use an interval logic based method, named the Duration Calculus (DC), to specify a simplified single-cell model of the DCA. Based on the DC specifications with further induction, we find that each individual cell in the DCA can perform its function as a detector in real-time. Since the DCA can be seen as many such cells operating in parallel, it is potentially capable of performing real-time detection. However, the analysis process of the standard DCA constricts its real-time capability. As a result, we conclude that the analysis process of the standard DCA should be replaced by a real-time analysis component, which can perform periodic analysis for the purpose of real-time detection.

## 1 Introduction

Artificial Immune Systems (AIS) [3] are computer systems inspired by both theoretical immunology and observed immune functions, principles and models, which can be applied to real world problems. As the natural immune system is evolved to protect the body from a wealth of invading micro-organisms, artificial immune systems are developed to provide the same defensive properties within a computing context. One of these immune inspired algorithms called the Dendritic Cell Algorithm (DCA) [6] is based on the function of the dendritic cells of the innate immune system. An abstract model of the behaviour of natural dendritic cells is used as the foundation of the developed algorithm. Currently, the DCA has been applied to numerous problems, including port scan detection [6], Botnet detection [1] and a classifier for robotic security [12]. They refer to the field of anomaly detection, which involves discriminating between normal and anomalous data, based on the knowledge of the normal data. The success of the applications has suggested that the DCA shows not only good performance on detection rate, but also the ability to reduce the rate of false alarms in comparison to other systems including Self Organising Maps [7]. However, one problem with DCA has been pointed out in [9], that is, the analysis process of the algorithm is performed offline rather than online in real-time. This results in the delays between when potential anomalies initially appear and when they are correctly identified. Such delays can be problematic for applications with strict time constraints, as they are often speed-critical. To solve this problem, it is desired to improve the real-time capability of the DCA, in order to develop an effective real-time detection system.

A real-time system [14] is a reactive system which, for certain inputs, has to compute the corresponding outputs within given time bounds (real-time criteria). The design of real-time systems generally requires high precision due to their particular application areas. The high precision is achieved by using formal methods that are based on the mathematical models of the systems being designed. The formal methods make it possible to specify the system properties at different levels and abstractions, as well as formally verify the specifications before implementing. One of the formal methods for specifying real-time systems is known as the Duration Calculus (DC) [17], which is a temporal logic and calculus for describing and reasoning about the properties of a real-time system over time intervals. The DC can specify the safety properties, bounded responses and duration properties of a real-time system, which can be logically verified through proper induction. Unlike predicate calculus [5] using time points to express time-depedent state variables or observables of the specified system, the DC uses time intervals with the focus on the implicit semantics level rather than the explicit syntactic level. As a result, it is more convenient and concise to use the DC to specify patterns or behaviour sequences of a real-time system over time intervals, compared to predicate calculus.

The real-time capability of the DCA should be assessed before making any improvement on the algorithm. In other words, it is essential to identify which properties of the algorithm can satisfy the real-time criteria, and which cannot. For this purpose, the DC is used to specify the behaviours of the DCA over particular time intervals. First of all, the DC specifications of the algorithm can be further induced by applying available proof rules in the DC. The mathematical aspects of the algorithm that have not been discovered might be revealed through the induction. In addition, the DC specifications include the temporal properties of the algorithm, which provide the insight of the duration required for each individual behaviour. The duration of each behaviour can be compared with the real-time criteria, to evaluate whether a behaviour can be performed in real-time or not. As a result, we can identify the properties of the algorithm that can satisfy the real-time criteria and those cannot at the behavioural level. The findings can be used as the basis for the further development of the real-time detection system based on the DCA.

The aim of this paper is to use the DC to specify the properties of the DCA, with the focus on the development of an effective real-time detection system. As a result, we would be able to identify the properties of the DCA that can be inherited for future development, and those need improved. Proper proof is included in this paper to support the conclusions derived from the DC specifications. The paper is organised as follows: the DCA is briefly described in section 2; the background information of the DC is given in section 3; the DC specifications of the single-cell model are given in section 4; the discussion of the analysis process of the DCA is provided in Section 5; finally the conclusions and future work are drawn in Section 6.

## 2 The Dendritic Cell Algorithm

### 2.1 Algorithm overview

As previously stated the blueprint for the DCA is the function of the dendritic cells of the innate immune system. Natural dendritic cells are capable of combining a multitude of molecular information and then interpret this information for the adaptive immune system, to induce appropriate immune responses towards perceived threats. Signal and antigen are the two types of molecular information processed by dendritic cells. Signals are collected from their local environment and consist of indicators of the health of the monitored tissue. Denrditic cells exist in one of three states of maturation to perform their immune function. In their initial immature state, dendritic cells are exposed to a combination of signals. They can differentiate into either semimature or fully mature state based on the concentrations of signals. Additionally, during their immature phase dendritic cells also collect debris in the tissue which are subsequently combined with the molecular environmental signals. Some of the debris collected are termed antigens, and are proteins originating from potential invading entities. Eventually dendritic cells combine evidence in the form of signals with the 'suspect' antigens to correctly instruct the adaptive immune system to respond, or become tolerant to the antigens in question. For more detailed information of natural dendritic cells, please refer to Lutz and Schuler [10].

The resulting algorithm incorporates the state transition pathway, the environmental signal processing procedure, and the correlation between signals and antigens. In the algorithm signals are represented as continuous real-number values and antigens are the categorical values of possible categories. The algorithm is based on a multi-agent framework, where each cell processes its own environmental signals and collects antigens. Diversity is generated within the cell population through the application of a 'migration threshold' - this value limits the number of signal instances an individual cell can process during its lifespan. This creates a variable time window effect, with different cells processing the signal and antigen input streams over a range of time periods [13]. The combination of signal/antigen correlation and the dynamics of a cell population are responsible for the detection capabilities of the DCA.

### 2.2 The single-cell model

The DCA consists of a population of artificial cells, each of which is capable of performing a set of identical behaviours, to accomplish its function as a detector.

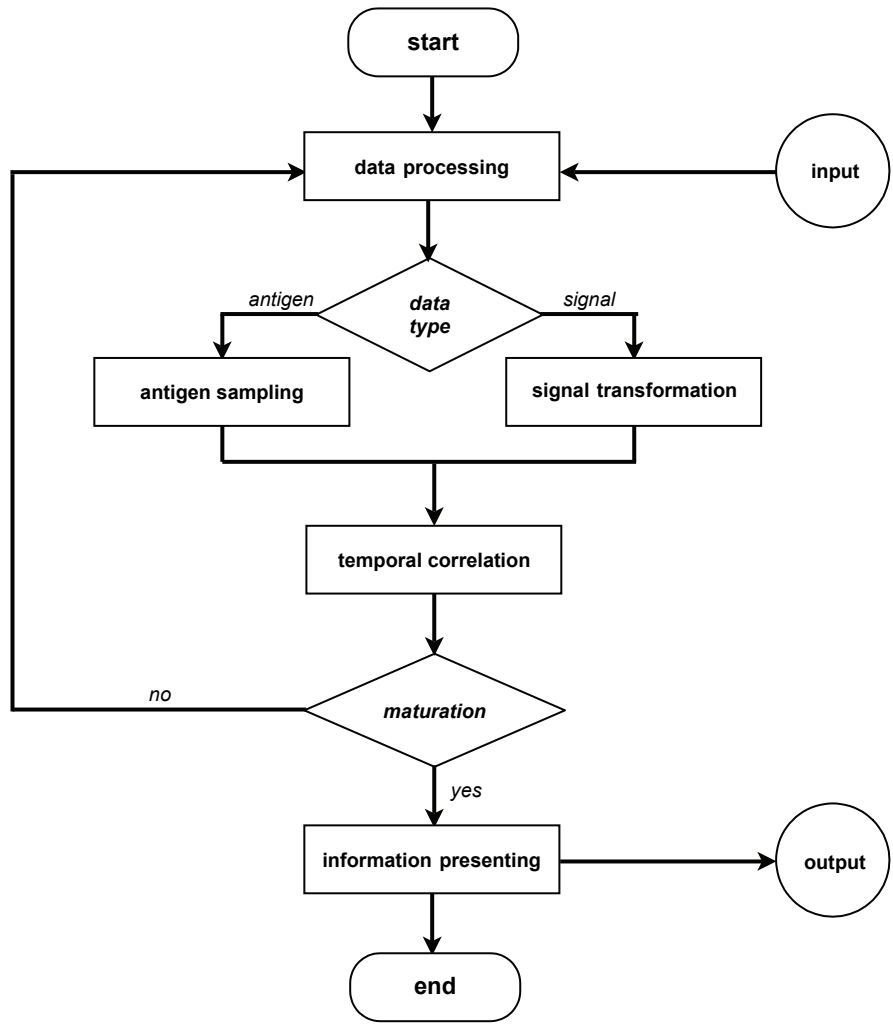

Fig. 1. The behavioural flowchart of the single-cell model

In order to understand the properties of the algorithm, we start with describing a simplified single-cell model. A one-cell model is demonstrated in [13] for the purpose of analysing the effect of signal frequency. Whereas, the single-cell model here is focused on the behavioural level of the DCA from a temporal perspective, rather than a quantitative level. The flowchart of the behaviours involved in the single-cell model is displayed in Fig. 1. The time-dependent behaviours of a cell are termed 'events' in this paper, and they are performed by the cell in each state during particular time intervals. The state and event mentioned in this paper are similar to those defined in temporal logic. Therefore, states must hold over any subintervals of an interval in which they hold, conversely events do not hold over any subintervals of an interval in which they hold. In other words, states can be broken down over multiple subintervals of an interval, whereas events cannot. The states, the events in each state, and the relevant time intervals of the single-cell model are included in the following description.

- **Immature state**: this is the initial state of the cell, where the cell is fed with input data instances. All the input data instances are handled by the data processing event, to determine their types. If the type of a data instance is 'signal', it is passed to the signals transformation event. Otherwise, if the type of data instance is 'antigen', the data instance is passed to the antigen sampling event. In each iteration of the system, only one signal instance but multiple antigen instances can be fed to the cell. The processed signals and sampled antigens are correlated by the temporal correlation event based on their time stamps. The cell keeps performing the events above cyclically, until the migration threshold is reached. This indicates that the cell has acquired sufficient information for decision making.

- **Matured state**: once the cell reaches its migration threshold, it changes from immature state to either semimature state or fully mature state. As same event is taken place in both semimature state and mature state, they are called 'matured state' in this paper. Based on the correlated signals and antigens by the temporal correlation event, the cell makes a decision on whether any potential anomalies appeared within the input data. Such decision is termed 'processed information' that is presented by the information presenting event as the output of the cell. Up to this point one lifespan of the cell is finished, and then the cell is reinitialised to immature state for new incoming data instances.

At the population level, the DCA can be seen as a systems in which multiple single-cell models are executed in parallel. The output of each matured cell is accumulated with the outputs of others in the population by the analysis process of the algorithm. From the accumulated outputs of all matured cells, the analysis process produces the final detection result in which the anomalies within the input data can be identified. In the standard DCA, the analysis process is performed after all instances of the input data are processed. This could make it difficult for the system to satisfy the real-time criteria if the size of the input data is large. The details will be discussed in section 5.

## 3   The Duration Calculus

The DC was firstly introduced by Zhou and Hansen [i6] as an extension of the Interval Temporal Logic [ii]. It uses continuous time for specifying desired properties of a real-time system without considering its implementation. The specifications are presented by DC formulas which express the behaviours of time-dependent variables or observables of a real-time system within certain time intervals. In DC specifications, not only abstract high-level but also detailed low-level specifications can be formulated according to the selected variables or observables. This makes it possible to specify the system from different perspectives at various levels. There are different versions of the DC [i7], including the classic DC, the extended DC and mean-value calculus. The work in this paper uses the classic DC, as it is sufficient for specifying the system presented.

In order to introduce the DC, the syntax defining the structure of DC specifications and the semantics explaining its meaning are described in this section. The DC specifications often consist of three elements, which are state assertions, terms and formulas. Their formal definitions given in [i4] are as following.

**Definition 1.** state assertions are Boolean combinations of basic properties of state variables, as defined in i.

$$P ::= 0 \mid 1 \mid X = d \mid \neg P \mid P_1 \wedge P_2 \tag{i}$$

As a state assertion, the Boolean value of the observable of $P$ can be either 0 or 1; It can have a state variable X whose value is $d$ of data type $D$; There are situations where $P$ does not hold; There are also situations where the substates of $P$, $P_1$ and $P_2$, both hold. The s e m a n t i c s of a state assertion involves the interpretation of time-dependent variables that occur within it. Let $\mathcal{I}$ be an interpretation, the semantics of a state assertion P is a function defined in 2.

$$\mathcal{I}\llbracket P \rrbracket : \mathsf{Time} \longrightarrow \{0, 1\} \tag{2}$$

where 0 or 1 represents the Boolean value of $P$ at $t \in$ Time, which can be also written as $\mathcal{I}(P)(t)$.

**Definition 2.** terms are expressions that denote real numbers related to time intervals, as defined in :

$$\theta ::= x \mid l \mid \int P \mid f(\theta_1, ..., \theta_n) \tag{3}$$

The expression above states that giving an interval $\int$ during which the state assertion $P$ holds, there is a global variable x that is related to the valuation a $n$-ary function $f$. The semantics of a term depends on the interpretation of state variables of the state assertion, the valuation of the global variables, and the given time interval. The semantics of a term $\theta$ is defined in 4.

$$\mathcal{I}\llbracket \theta \rrbracket : \mathsf{Val} \times \mathsf{Intv} \longrightarrow \mathbb{R} \tag{4}$$

where Val stands for the valuation $(\mathcal{V})$ of the global variables, and Intvl is the given interval which can be defined in 5.

$$\mathsf{Intv} \stackrel{\text{def}}{\Longleftrightarrow} \{[b,e] \mid b,e \in \mathsf{Time} \text{ and } b \le e\} \tag{5}$$

So this term can also be written as $\mathcal{I}[\![\theta]\!](\mathcal{V},[b,e])$. υ (ν, [b, e]).

**Definition 3.** *formulas* describe properties of obervables depending on time intervals, as defined in 6

$$F ::= p(\theta_1, ..., \theta_n) \mid \neg F_1 \mid F_1 \wedge F_2 \mid \forall x \bullet F_1 \mid F_1 \; ; \; F_2 \tag{6}$$

This expression shows that there is a $n$-ary predicate with the terms of $\theta_1, ..., \theta_n$ defined in the interval of $\int$, during which $F_1$ does not hold or $F_1$ and $F_2$ hold.

The quantitative part of the expression is separated by the symbol of ' $\bullet$ '. It states that for all $x$, $F_1$ holds in the interval of $\int$, or there are situations where $F_1$ and $F_2$ hold respectively in the subintervals of $\int$. The symbol ';' is the chop operator used for dividing the given time interval into subintervals. The semantics of a formula involves an interpretation of the state variables, a valuation of the global variables and a given time interval, defined in 7. The relevant state variables and global variables all appear in the terms of this formula.

$$\mathcal{I}[\![F]\!] : \mathsf{Val} \times \mathsf{Intv} \longrightarrow \{\mathsf{tt}, \mathsf{ff}\} \tag{7}$$

where **tt** stands for true and **ff** for false. It can also be written as $\mathcal{I}[\![F]\!](\mathcal{V}, [b,e])$, which stands for the truth value of $F$ under the interpretation $\mathcal{I}$, the valuation $\mathcal{V}$, and the interval $[b, e]$.

## 4  DC specifications of the system

Before going into the details of the DC specifications, we want to introduce the notations that are used in this section, listed as following.

- $I : \mathsf{Time} \longrightarrow \{0,1\}$ is the Boolean observable indicating that the cell is in immature state.
- $M : \mathsf{Time} \longrightarrow \{0,1\}$ is the Boolean observable indicating that the cell is in matured state.
- $E_i : \mathsf{Time} \longrightarrow \{0,1\}$ is the Boolean observable representing the $i$th event is being performed, where $i \in \mathbb{N}$.
- $l_i \in \mathbb{R}$ is the duration time of $E_i$.
- $l_a \in \mathbb{R}$ is the duration time of the analysis process.
- $b \in \mathbb{R}$ is the real-time bound, if a processed is completed within $b$, then it can be performed in real-time, and vice versa.
- $r \in \mathbb{R}$ is the duration of one iteration in the system.
- $c \in \mathbb{R}$ is the duration of one lifespan during which the cell experiences both immature state and matured state.

– $\bar{m} \in \mathbb{N}$ is the average number of processed signal instances within one lifespan of the cell.
– $\bar{n} \in \mathbb{N}$ is the average number of sampled antigen instances within one lifespan of the cell.

To be more specific, the definition of each event $E_i$ is shown as follows.

– $E_1$ is the *data processing* event with an interval $l_1$.
– $E_2$ is the *signal transformation* event with an interval $l_2$.
– $E_3$ is the *antigen sampling* event with an interval $l_3$.
– $E_4$ is the *temporal correlation* event with an interval $l_4$.
– $E_5$ is the *information presenting* event with an interval $l_5$.

## 4.1 Specifications of the single-cell model

According to the description of the single-cell model in section 2, the cell performs a set of particular events in each state. So the states of a cell can be indicated by the combination of whether $E_i$ is being performed (the Boolean observable of $E_i$), as shown in 8.

$$I ::= 0 \mid 1 \mid \neg I \mid E_1 \vee (E_2 \wedge \neg E_3) \vee (\neg E_2 \wedge E_3) \vee E_4$$
$$M ::= 0 \mid 1 \mid \neg M \mid E_5 \tag{8}$$

In immature state (1), the cell is fed with input data instances whose type can be either signal or antigen. The immature state can be indicated by $E_1$ holds, $E_2$ $\wedge$ $\neg E_3$ holds, $\neg E_2$ $\wedge$ $E_3$ holds, or $E_4$ holds. Conversely, in matured state (M), the cell presents the processed information from correlated signals and antigens. The matured state can be indicated by $E_5$ holds.

The specifications in 8 can be expanded by including the time interval of each event, expressed in the form of formulas, in which the temporal dependencies between events are included. For example, $E_2$ or $E_3$ depends on the completion of $E_1$, and only either of them can be performed at one point; $E_4$ depends on the completion of $E_2$ and $E_3$, as the temporal correlation requires both processed signals and sampled antigens; $E_5$ is performed as soon as the cell changes to matured state, it is not dependent on any other events. Two formulas that are corresponding to the immature state and matured state of a cell are shown in 9.

$$F_1 ::= \rceil[I] \mid \neg E_5 \mid (E_1 \ ; \ E_2 \wedge \neg E_3) \ ; \ (E_1 \ ; \ \neg E_2 \wedge E_3) \ ; \ E_4$$
$$F_2 ::= \lceil M \rceil \mid \neg(E_1 \vee E_2 \vee E_3 \vee E_4) \mid E_5 \tag{9}$$

where $\lceil I \rceil$ stands for that $I$ holds almost everywhere within the time interval constrained by formula $F_1$, and $\lceil M \rceil$ stands for that M holds almost everywhere within the time interval constrained by $F_2$. So in the interval constrained by $F_1$, it is certain that $E_5$ does not hold. This interval can be divided into multiple subintervals in which $E_1$, $E_2$ $\wedge \neg E_3$, $\neg E_2 \wedge E_3$, or $E_4$ holds respectively. In the interval constrained by $F_2$, none of $E_1$, $E_2$, $E_3$ or $E_4$ holds, but only $E_5$ holds.
So for instance, the overall length of the time interval while $F_1$ and $F_2$ holds is six, and the time interval of each event is equal to one, Fig. 2 shows the interpretation of the two formulas.

As the cell can process multiple signal instances and antigen instances before it gets matured, some of the events can be performed for more than once within one lifespan of the cell. To generalise this, the average numbers ($\bar{m}$ and $\bar{n}$) of processed data instances are used. Therefore, the pattern of '$E_1$ ; $E_2 \wedge \neg E_3$' appears $\bar{m}$ times, and the pattern of '$E_1$ ; $\neg E_2 \wedge E_3$' appears $\bar{n}$ times. Additionally, $E_4$ is performed $\bar{m}$ times, as the number of performed temporal correlations is equvalent to the number of prossed signal instances. However, $E_5$ is only performed once, as it is irrelevant to the number of processed signal instances or sampled antigen instances. As a result, the duration of a cell being in immature state and the duration of a cell being in matured state can be formalised as in 10. The duration of one lifespan of the cell is defined as $C = \int I + \int M$.

$$
\begin{aligned}
\int I &= \bar{m} \cdot (l_1 + l_2) + \bar{n} \cdot (l_1 + l_3) + \bar{m} \cdot l_4 \\
\int M &= l_5
\end{aligned}
\tag{10}
$$

## 4.2   Evaluation of the real-time capabability

Based on the DC specifications above, we conduct a test to examine the real-time capabability of an individual cell of the DCA. If the cell completes at least one cell cycle within the given real-time bound ($($), it suggests that the cell can perform its function in real-time. This test is formalised as a requirement in 11.

$$
\mathsf{Req} \stackrel{\text{def}}{\iff} \square (b \geq (\bar{m}+1) \cdot r \implies \int I + \int M \leq b)
\tag{11}
$$

where E is the dual modal operator of interval logic, defined as EF holds in an interval of $[($, e$]$ only if F holds in every subinterval of $[($, e$]$. The condition of the Req is the left side of the logical connective '$\Longrightarrow$', while the conclusion is on the right side. If this requirement is satisfied, then we conclude that each individual cell in the DCA is capable of operating in real-time.

As mentioned in section 2, a cell exists in either immature state or matured state, and all the events within each state should be performed in each iteration. According to the definition, one system iteration is equal to the duration between two successive updates of signal instance. In each iteration the cell processes one signal instance but a number of antigen instances. Therefore, the cumulative duration of $E_1$, $E_2$ and $E_4$ should not be greater than the duration of one iteration, and the duration of $E_5$ should also not be greater than the duration of one iteration. Such properties can be formalised as two design decisions of the single-cell model shown in 12.

$$
\begin{aligned}
\mathsf{Des\text{-}1} &\stackrel{\text{def}}{\iff} \square(\lceil I \rceil \implies l_1 + l_2 + l_4 \leq r) \\
\mathsf{Des\text{-}2} &\stackrel{\text{def}}{\iff} \square(\lceil M \rceil \implies l_5 \leq r)
\end{aligned}
\tag{12}
$$

The two design decisions are the extra preconditions that determine whether the system can satisfy the real-time criteria or not, as defined in **Theorem 1**.

**Theorem 1.** $\models$ (Des-1 $\land$ Des-2) $\longrightarrow$ Req
It expresses that if both design decisions Des-i and Des-2 hold, the requirement Req can be satisfied.

**Proof:**

$$b \geq (\bar{m} + 1) \cdot r$$
$\Longrightarrow$ {the cell exists in immature state or matured state}
$$\lceil I \rceil \,;\, \lceil M \rceil$$
$\Longrightarrow$ {by formula 9}
$$\left(\int I\right) \,;\, \left(\int M\right)$$
$\Longrightarrow$ {by formula 10}
$$\left(\int I = \bar{m} \cdot (l_1 + l_2) + \bar{n} \cdot (l_1 + l_3) + \bar{m} \cdot l_4\right) \,;\, \left(\int M = l_5\right)$$
$\Longrightarrow \left(\int I = \bar{m}(l_1 + l_2 + l_4) + \bar{n}(l_1 + l_3)\right) \,;\, \left(\int M = l_5\right)$
$\Longrightarrow$ {by **Des-1** and **Des-2**}
$$\left(\int I \leq \bar{m} \cdot r\right) \,;\, \left(\int M \leq r\right)$$
$\Longrightarrow$ {by the addition rule of calculus}
$$\int I + \int M \leq \bar{m} \cdot r + r = (\bar{m} + 1) \cdot r$$
$\Longrightarrow \int I + \int M \leq b$

Thus Req holds on every interval of $\ell \geq (\bar{m} + 1) \cdot r$, and **Theorem 1** is proved. As the increase of iterations is not affected by the events for processing the antigen instances, the duration of these events is eliminated in the induction above.

Based on **Theorem 1**, as long as the real-time bound is not smaller than the duration of '$(\bar{m} + 1) \cdot r$', the cell can at least complete one lifespan. According to the experiments performed in [9], the value of $\bar{m}$ is normally smaller than 10. Therefore, the single-cell model can satisfy the real-time criteria if the real-time bound is not less than the duration of 11 iterations, which can be easily satisfied in most applications. This suggests that a single cell in the DCA can be performed in real-time. As a consequence, the DCA can perform all the events except the analysis process in real-time, since the algorithm employs a population of such cells that operate in parallel.

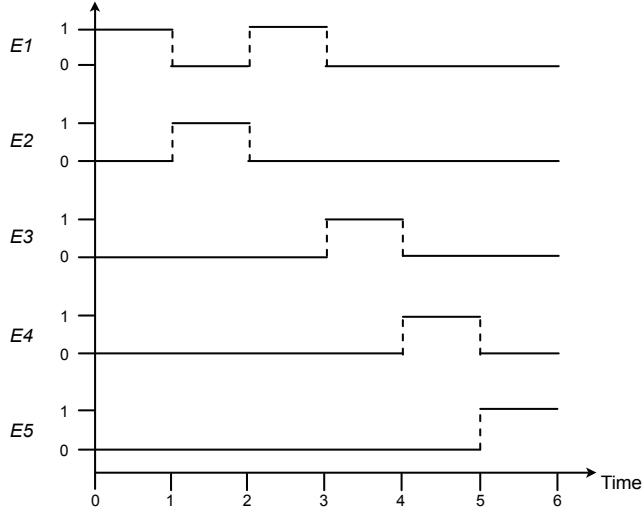

Fig. 2. Interpretation for $E_1$, $E_2$, $E_3$, $E_4$, and $E_5$, and the whole interval is divided into subintervals by the events.

## 5   Discussion of the analysis process

The DCA also involves an analysis process that produces the final detection result from the accumulated outputs of matured cells in the population. As mentioned before, the analysis process of the standard DCA is performed offline after all the instances of the input data are processed. In the case that the input data consist of m (m ∈ N) signal instances, by formula 10, $\frac{m}{\overline{m}}$ lifespans of the cell are required to process the whole input data. To satisfy the real-time bound, the duration needed for the standard DCA to get the final detection result can be formalised as in 13.

$$C \cdot \frac{m}{\overline{m}} + l_a \leq ($$ (13)

As $C$, $\overline{m}$ and $l_a$ are constants, whether formula 13 can hold or not is determined by the quantity of m. The value of m is derived from the number of signal instances contained within the input data. As the size of the input data grows, the number of signal instances is getting bigger and bigger. This can cause that the duration for getting the final detection result increases dramatically and exceeds the real-time bound. Therefore, as the size of the input dataset increases, it is becoming more and more difficult for the standard DCA to satisfy the real-time criteria. Therefore, the analysis process of the standard DCA is the weakness of the algorithm in terms of real-time detection.

In order to satisfy the real-time criteria, the analysis process of the standard DCA should be replaced by a real-time analysis component that performs periodic analysis during detection. This can be achieved by segmenting the current output of the DCA, which is performed in a variety of ways, as suggested in [8]. Segmentation involves slicing the output data of the DCA into smaller segments with a view to generating finer grained results and to perform analysis in parallel with the detection process. Segmentation can be performed based on a fixed quantity of output data items or alternatively on a basis of a fixed time period. As the analysis process is performed within each segment, the sub-duration for each segment to produce the detection result is much shorter than the whole duration. It hightly possbile for the sub-duration to satisfy the real-time bound. Moreover, such sub-duration can be made to satisfy the real-time bound, by adjusting the segment size that determines the length of each sub-duration. Segmentation is the initial step of developing the real-time analysis component, and eventually an approach that can deal with the online dynamics is required. This approach should be able to adapt and evolve during detection, so that it can deal with the new situations that have not been previously seen. This leads to the future work of dynamic segmentation.

## 6    Conclusions and future work

In this paper, we used the DC specifications to formally describe a simplified single-cell model of the DCA. The temporal properties of the events performed by the cell in each state are included, indicating the dependencies between events. To explore the real-time capabability of the DCA, we conducted a test from which Theorem 1 is derived. The conclusion of Theorem 1 suggests that each cell of the DCA can operate in real-time, based on the single-cell model. As the DCA employs a population of such cells that operate in parallel, the events functioning detection can be performed in real-time. Therefore, the DCA is potentially capable of performing real-time detection. However, the analysis process of the standard DCA is performed after all the instances of the input data are processed. As a result, if the size of input dataset grows, the duration required for the algorithm to produce the final detection result increases dramatically. This make it more and more difficult for the algorithm to satisfy the real-time criteria. Therefore, in order to develop an effective real-time detection system based on the DCA, the analysis process of the algorithm needs to be replaced by a real-time analysis component, which is capable of performing periodic analysis during detection. Preliminary work on segmentation has been done in [8], and the result appears promising. Eventually an adaptive real-time analysis component that incorporates with dynamic segmentation will be developed.

The DC specifications in this paper focus on the behavioural level of the single-cell model without going into any further details, which is sufficient for the scope of the paper. However, for future work the population level of DCA should also be covered to better present the algorithm. In addition, the DC is mainly used for specifying the requirement of real-time systems, to design and implement real-time systems, other formal methods are also required. These methods include the Timed Automata [2] and the PLC Automata [4], which can be used for modelling cyclic behaviours of interacting objects in real-time systems. Therefore, they are ideal for formally modelling the systems like the DCA that is based on multi-agent framework. Moreover, there are existing tools, such as UPPAAL [15] and so on, which facilitate the automatic verification of the systems modelled in the Timed Automata and PLC Automata. As a result, the designed real-time system can be formally verified before its implementation.